\documentclass[conference,hidelinks]{IEEEtran}
\IEEEoverridecommandlockouts
\usepackage{cite}
\usepackage{amsmath,amssymb,amsfonts}
\usepackage{array}
\usepackage[caption=false,font=small,labelfont=sf,textfont=sf]{subfig}
\usepackage{textcomp}
\usepackage{stfloats}
\usepackage{url}
\usepackage{graphicx}
\usepackage{algorithm}
\usepackage{algpseudocode}
\usepackage{makecell,booktabs}
\usepackage{multirow}
\usepackage{siunitx}
\usepackage{xcolor}
\usepackage{enumitem}
\usepackage{xspace}
\usepackage{tcolorbox}
\usepackage{inconsolata} % 更好看的等宽字体（可选）
\usepackage{orcidlink}
\usepackage{colortbl}

\definecolor{headerblue}{RGB}{222,235,247}
\definecolor{blockgray}{RGB}{242,242,242}

\usepackage{booktabs}
\usepackage{tabularx}
\usepackage{ragged2e}

\usepackage{soul, color,}

\usepackage{pgfplots}
\pgfplotsset{compat=1.18}

\providecommand{\Description}[2][]{}

\def\BibTeX{{\rm B\kern-.05em{\sc i\kern-.025em b}\kern-.08em
    T\kern-.1667em\lower.7ex\hbox{E}\kern-.125emX}}
\usepackage{balance}
\begin{document}

\title{Category-Anchored Large Language Models for Next POI Recommendation
% {\footnotesize \textsuperscript{*}Note: Sub-titles are not captured for https://ieeexplore.ieee.org  and
}

% \author{\IEEEauthorblockN{Lexi Bai\textsuperscript{\orcidlink{0009-0003-2260-9079}}}
% \IEEEauthorblockA{\textit{School of Computer Science and Technology} \\
% \textit{Harbin Institute of Technology}\\
% Shenzhen, China \\
% bailexi@stu.hit.edu.cn}
% \and
% \IEEEauthorblockN{2\textsuperscript{nd} Xiaopeng~Zeng\orcidlink{0009-0006-4826-7187}}
% \IEEEauthorblockA{\textit{School of Computer Science and Technology} \\
% \textit{Harbin Institute of Technology}\\
% Shenzhen, China \\
% xpzeng666@gmail.com}
% \and
% \IEEEauthorblockA{\textit{ School of Computer Science and Technology} \\
% \textit{ Harbin Institute of Technology}\\
% Shenzhen, China \\
% \and
% \and
% \IEEEauthorblockA{\textit{ School of Computer Science and Technology} \\
% \textit{ Harbin Institute of Technology}\\
% Shenzhen, China \\
% \and

\author{
\IEEEauthorblockN{Lexi~Bai\textsuperscript{1}\orcidlink{0009-0006-4826-7187},~Xiaopeng~Zeng\textsuperscript{1}\orcidlink{0009-0004-6884-9083},~Xun~Zhou\textsuperscript{1}\orcidlink{0000-0003-4930-6572},~Yuanshao~Zhu\textsuperscript{2,*}\orcidlink{0000-0002-5657-181X},~and James~Jianqiao~Yu\textsuperscript{1,*}\orcidlink{0000-0002-6392-6711}}\thanks{*Corresponding author}
    % duplicate the line above as many times as needed to list all authors
    %%%% author affiliations
    \IEEEauthorblockA{\textsuperscript{1}Harbin Institute of Technology, Shenzhen, China \\
    \textsuperscript{2}City University of Hong Kong, Hong Kong SAR, China.}
    % duplicate the line above as many times as needed to list all affiliations
    %%%% corresponding author contact details
    \IEEEauthorblockA{bailexi@stu.hit.edu.cn,~xpzeng666@gmail.com,~zhouxun2023@hit.edu.cn,~yuanshao@ieee.org, ~jqyu@ieee.org}

}

\maketitle

\begin{abstract}
Next point-of-interest (POI) recommendation aims to predict a user's next destination from spatio-temporal check-in data, and serves as the foundation for many location-based services.
Recent advances incorporate large language models (LLMs) into this task by textualizing user trajectories to exploit their semantic reasoning ability.
However, directly generating POI identifiers with such models is unreliable: POIs are sparse, localized, and long-tailed entities with weak natural-language semantics, which creates a semantic-to-entity gap between high-level mobility intent and concrete POI prediction.
To bridge this gap, we propose ANCHOR, a category-anchored framework for semantic-to-entity grounding in next POI recommendation.
Specifically, ANCHOR first estimates category-level mobility intent in a compact and stable semantic space, rather than predicting concrete POI entities directly.
The inferred intent is then grounded into a candidate set by integrating three complementary sources of mobility evidence: category anchors, historical visit memory, and user preferences.
Finally, a history-aware reranker selects the target POI by matching each candidate with the user's historical mobility patterns and current context.
Experiments on three real-world check-in datasets show that ANCHOR consistently outperforms representative traditional, structure-enhanced, and LLM-based baselines on the reported metrics.
Further analyses validate the effectiveness of semantic-to-entity grounding and show favorable trade-offs among candidate coverage, ranking quality, and inference efficiency.
% Recent advances attempt to incorporate large language models (LLMs) into this task by textualizing user trajectories and leveraging their semantic reasoning ability.
% Despite their promise, directly generating POI identifiers with LLMs is unreliable, since POIs are sparse, localized, and long-tailed entities with weak natural-language semantics.
% This creates a semantic-to-entity gap between high-level mobility intent and concrete POI prediction.
% We propose ANCHOR, a category-anchored framework that bridges this gap through semantic-to-entity grounding.
% Instead of asking the LLM to predict POI entities directly, ANCHOR first estimates category-level mobility intent in a compact semantic space.
% It then grounds the inferred intent into a candidate POI set by integrating category anchors, history POI memory, and user preference evidence, followed by a history-aware reranker that selects the final POI under the target context.
% Experiments on three real-world check-in datasets show that ANCHOR consistently outperforms representative traditional, structure-enhanced, and LLM-based baselines.
% Further analyses show that ANCHOR is robust across different history sparsity settings and achieves favorable trade-offs between candidate coverage, ranking quality, and efficiency.
\end{abstract}


\section{Introduction}\label{sec:intro}
Driven by the ubiquity of location-based services and urban sensing systems, massive user mobility traces are continuously recorded as spatio-temporal check-in data~\cite{cho2011friendship}.
Over such data, next point-of-interest (POI) recommendation aims to infer a user's next concrete destination from historical visits, supporting a wide range of urban analytics and location-aware services~\cite{cheng2013you,feng2015personalized}.
This task is challenging because it requires reasoning over sparse and heterogeneous mobility traces while capturing complex spatio-temporal dependencies and sequential movement patterns~\cite{liu2016predicting,lian2020geography,luo2021stan,yang2022getnext}.
Consequently, effective trajectory modeling for reliable next POI recommendation remains an important problem in spatio-temporal data analysis.

To address this task, existing studies have explored next POI prediction from multiple angles.
Early efforts show that user preference and geographical regularity are important signals for modeling human mobility over historical check-ins~\cite{cheng2013you,feng2015personalized}.
Subsequent studies cast the problem as sequential or spatio-temporal trajectory prediction, where neural models capture dynamic transitions and historical mobility dependencies in user trajectories~\cite{liu2016predicting,feng2018deepmove,yang2020location,luo2021stan}.
Recent work on human mobility prediction further highlights the importance of adapting mobility models to changing spatio-temporal patterns~\cite{han2025adamove}.
Beyond individual trajectories, recent methods also incorporate collaborative and structural signals in check-in data.
GETNext constructs trajectory flow maps to transfer mobility patterns across users~\cite{yang2022getnext}, while STHGCN models high-order spatio-temporal relations with hypergraphs~\cite{yan2023spatio}.
More recently, large language models (LLMs) have been introduced into mobility modeling by converting trajectories, contexts, or POI semantics into language-readable forms, so as to leverage semantic knowledge and contextual reasoning for prediction~\cite{li2024large,wang2025generative}.
Overall, these efforts reveal a clear trend from modeling discrete check-in patterns toward integrating spatio-temporal structure, collaborative signals, and semantic knowledge.

\begin{figure}[t]
\centering
\includegraphics[width=\columnwidth]{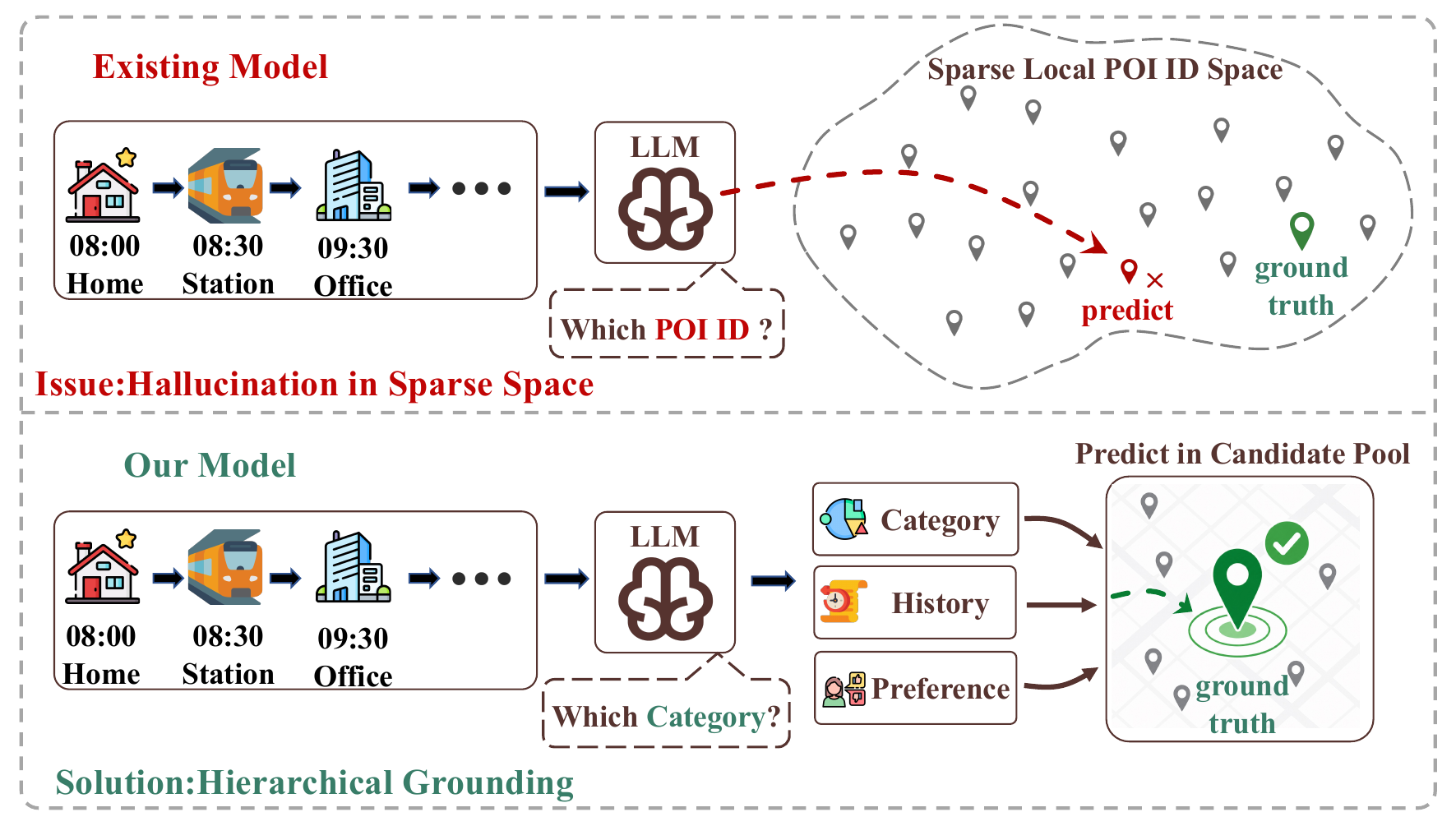}
\caption{Comparison between existing LLM-based POI recommendation models and ANCHOR.
Existing models directly ask the LLM to generate a concrete POI ID from user history, which is unreliable in sparse local POI ID spaces.
ANCHOR first predicts category-level intent and then grounds it into concrete candidate POIs using category, history, and preference evidence.}
\label{fig:intro_motivation}
\end{figure}

Despite this progress, a critical gap remains.
Existing methods mainly make predictions over the concrete POI entity space, where the target is a localized place rather than an abstract semantic concept.
Sequential, graph-based, and spatio-temporal models can capture transition regularities from historical trajectories, but they usually lack an explicit mechanism to model the high-level mobility intent behind the next visit.
LLM-based methods, by contrast, can reason over trajectory semantics and infer possible visiting purposes from textualized mobility context.
However, next POI recommendation ultimately requires selecting a concrete POI from a sparse, localized, and dataset-specific entity space.
Because POI identifiers and names often provide weak and unstable semantics, directly generating the target POI may lead to entity mismatch, sparse supervision, and unreliable generation.
We refer to this granularity mismatch between semantic intent and concrete entity prediction as the semantic-to-entity gap.

To bridge this gap, an effective prediction framework should go beyond textualizing trajectories and directly generating POI IDs.
Instead, it should decompose next POI prediction into three connected steps:
(i) \textbf{reasoning} about the user's next move in a compact and stable semantic space;
(ii) \textbf{grounding} the inferred intent into concrete and comparable candidate POIs;
and (iii) \textbf{discriminating} among these candidates using historical and contextual evidence.
Together, these steps form a reason--ground--discriminate pipeline that turns high-level mobility intent into entity-level POI prediction.
As illustrated in Fig.~\ref{fig:intro_motivation}, our design contrasts with direct POI generation over a sparse and localized entity space.

However, realizing this pipeline is non-trivial, because check-in data introduce three concrete challenges.
\noindent\textbf{C1: Symbolic POI Space.}
In real-world check-in data, POIs are usually represented as dataset-specific IDs or local entity identifiers rather than natural-language objects with stable semantics.
This makes the raw POI space difficult for LLMs to interpret, compare and generate reliably.
\noindent\textbf{C2: Coarse Intent Ambiguity.}
Even when high-level mobility intent is correctly inferred, it may correspond to many concrete POIs with different historical relevance, spatial proximity, visiting time, and transition patterns.
This creates an inherent one-to-many ambiguity from semantic intent to entity candidates.
\noindent\textbf{C3: Contextual Evidence Conflict.}
Reliable POI prediction depends on heterogeneous evidence from user history, temporal context, spatial proximity, and structural relations.
However, the importance of these signals varies across users and scenarios: a frequently visited POI may be implausible at the current time, while a structurally related POI may still be spatially distant.
Such context-dependent conflicts make it difficult to determine which evidence should dominate the final prediction.

To address these challenges, we propose ANCHOR, a category-anchored semantic-to-entity grounding framework for next POI recommendation over spatio-temporal check-in data.
Specifically, ANCHOR first leverages an LLM to estimate the user's high-level mobility intent in a more stable category space.
It then grounds the inferred intent into a concrete POI candidate set by combining category-based recall with historical mobility evidence and personalized preference signals.
This multi-source grounding process alleviates the one-to-many ambiguity from coarse intent to concrete POIs while preserving semantic constraints.
Finally, ANCHOR performs history-aware reranking to distinguish grounded candidates according to their compatibility with the user's historical mobility patterns.
By connecting LLM-based semantic reasoning with structured POI entity prediction, ANCHOR provides a grounded prediction pipeline for next POI recommendation over sparse and localized mobility data.

Our contributions can be summarized as follows:
\begin{itemize}[leftmargin=*]
\item We identify the semantic-to-entity gap in next POI recommendation over spatio-temporal check-in data, and propose ANCHOR, which uses category-level intent as a semantic anchor to bridge high-level mobility intent and concrete POI prediction.

\item We design a multi-source candidate grounding and history-aware reranking framework that transforms coarse semantic intent into concrete POI candidates from complementary sources and adaptively calibrates them under the target context.

\item Extensive experiments on three real-world datasets show that ANCHOR consistently achieves state-of-the-art performance over various baseline models, while maintaining favorable trade-offs among candidate coverage, ranking quality, and inference efficiency.
\end{itemize}

\section{Preliminaries}

\noindent\textbf{Definition 1 (Check-in).}
Let $\mathcal{U}$, $\mathcal{P}$, and $\mathcal{C}$ denote the sets of users, POIs, and POI categories, respectively.
A check-in record can be denoted as $q=(u,~p,~c,~t,~g)$, which means user $u\in\mathcal{U}$ visits POI $p\in\mathcal{P}$ at timestamp $t$.
Here, $c\in\mathcal{C}$ denotes the category of POI $p$, and $g=(\text{lat},~\text{lon})$ denotes the corresponding geographical coordinate.

\noindent\textbf{Definition 2 (User Trajectory).}
For a user $u$, a trajectory is a time-ordered sequence of check-ins, denoted as $T_u=\langle q_1,~q_2,\ldots,~q_n\rangle$.
Each check-in $q_i=(u,~p_i,~c_i,~t_i,~g_i)$ records the visited POI, POI category, timestamp, and geographical coordinate.
The timestamps satisfy $t_i<t_{i+1}$.

% \noindent\textbf{Definition 3 (Candidate POI Set).}
% Given a user trajectory $T_u$ and a target timestamp $t_{n+1}$, a candidate POI set is denoted as $\mathcal{P}_{\text{cand}}\subseteq\mathcal{P}$.
% It contains the POIs to be further ranked for next-POI recommendation.

\noindent\textbf{Problem Definition.}
For a user $u$ and the user's historical trajectory $T_u$. 
The next POI recommendation aims to predict the next POI $p_{n+1}\in\mathcal{P}$ that the user will visit for the target timestamp $t_{n+1}$.
Specifically, the model outputs a ranked list of candidate POIs:
\begin{equation}
\hat{\mathcal{Y}}_{u}
=
% \operatorname*{arg\,topK}_{p\in\mathcal{P}}
\operatorname*{TopK}_{p\in\mathcal{P}}
s(p\mid T_u,t_{n+1}),
\end{equation}
where $s(\cdot\mid\cdot)$ denotes the predicted preference score and $\hat{\mathcal{Y}}_{u}$ serves as the final prediction with top-ranked.

\section{Methodology}

\subsection{Framework Overview}
\begin{figure*}[t]
\centering
\includegraphics[width=\textwidth]{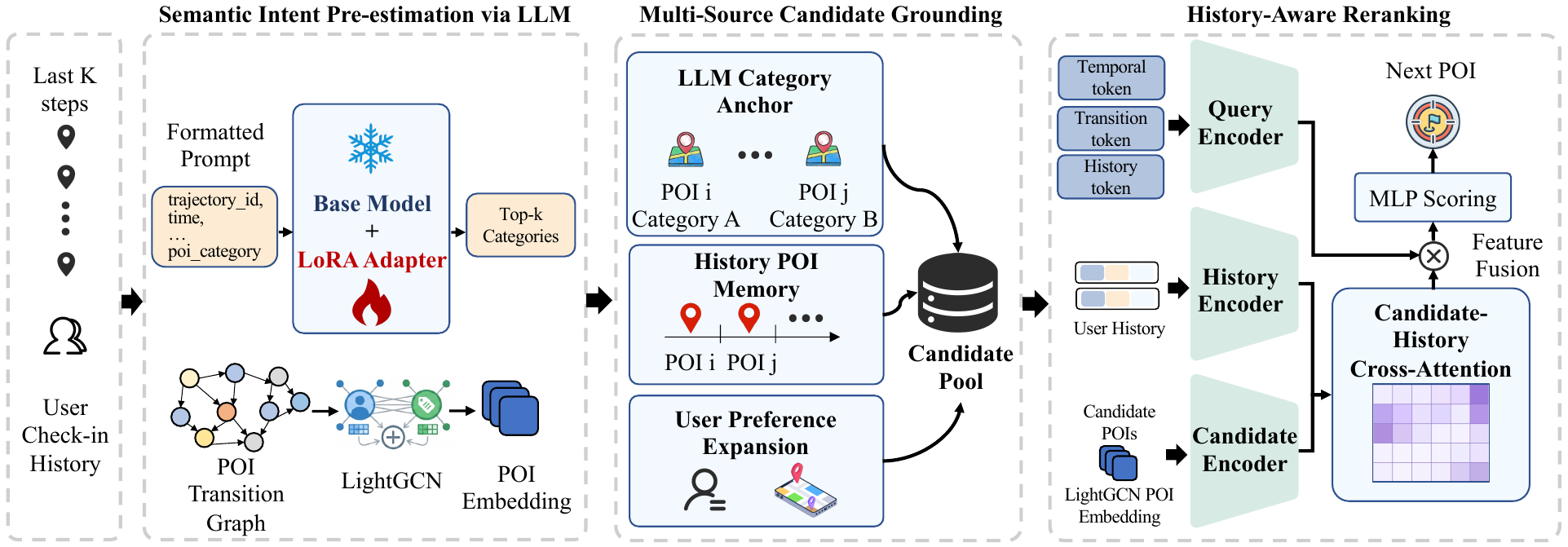}
\caption{Overview of the proposed ANCHOR framework, which consists of three modules: semantic intent pre-estimation for predicting category-level intent, multi-source candidate grounding for constructing the candidate POI pool, and history-aware reranking for producing the final recommendation.}
\label{fig:framework}
\end{figure*}

As shown in Fig.~\ref{fig:framework}, ANCHOR decomposes LLM-based next POI recommendation into three consecutive modules: semantic intent pre-estimation, multi-source candidate grounding, and history-aware reranking, which respectively address the three challenges identified in Section~\ref{sec:intro}.
\begin{itemize}[leftmargin=*] 
\item \textbf{Semantic Intent Pre-estimation.} Given the user's visible check-in history, ANCHOR uses the LLM to infer high-level mobility intent in a compact and stable category space, rather than generating a concrete POI ID.

\item \textbf{Multi-source Candidate Grounding.} Taking the top-$K_c$ predicted categories as semantic anchors, ANCHOR grounds the coarse category intent into a concrete candidate POI set by combining category anchors with the user's historical visits and personalized preferences.

\item \textbf{History-aware Reranking.} ANCHOR reranks the candidates by modeling their compatibility with the user's historical behavior and the target context, adaptively weighting heterogeneous evidence for the final prediction.
\end{itemize}
Through this decomposed design, ANCHOR bridges the gap between high-level mobility intent and concrete POI entities, transforming the semantic understanding of LLMs into reliable POI-level recommendations.

\subsection{Semantic Intent Pre-estimation via LLM}

POIs in the real world are typically represented as dataset-specific symbolic identifiers drawn from a sparse and long-tailed space, which makes them difficult for an LLM to generate or classify reliably. 
In contrast, POI categories capture the high-level semantic intent behind a visit (such as dining, shopping, and transportation) forms a smaller, semantically stable space.
In practice, such categories are better matches for the language understanding ability of LLMs. 
Therefore, we first use the LLM to estimate category-level mobility intent, which comprises three components: trajectory prompting, an LLM category classifier, and LoRA fine-tuning.

\noindent\textbf{Trajectory Prompting.}
To enable an LLM to infer user intent from check-in history, the first step involves converting the raw location data into a prompt that the model can interpret.
A naive approach is to expand each check-in into a complete natural-language sentence. 
However, recent studies show that this causes the prompt length to grow rapidly with history size, limiting how much historical behavior can fit within the input context~\cite{li2024large}. 
To avoid this overhead, ANCHOR adopts a compact \emph{field-based} prompting strategy that serializes each check-in into a concise sequence of spatio-temporal attributes separated by vertical bars.

Specifically, given the visible history $T_u$ before the target time $t$, ANCHOR constructs a trajectory prompt:
\begin{equation}
  X_u=\mathrm{Prompt}(T_u, u, t). 
\end{equation}
As shown in Table~\ref{tab:trajectory_prompt}, the prompt is organized into four functional blocks: an instruction block, a history block, a current block, and an output block. 
Here, the history block records past visits exactly as \texttt{trajectory\_id | time | day\_of\_week | norm\_in\_day\_time | poi\_category}. 
\texttt{norm\_in\_day\_time} denotes the visiting time normalized within a day. 
For the target check-in, the current block provides the target trajectory identifier and its temporal context, but masks the target category. 
Finally, the output block requires the model to predict the category of the target visit from the provided context.

By emphasizing semantic categories and temporal patterns instead of concrete POI identifiers, this design enables the LLM to better infer the user's underlying mobility intent.
More importantly, by employing this compact representation, ANCHOR significantly reduces token overhead compared with sentence-level textualization (see Section~\ref{sec:exp_efficiency}).

% To mitigate this issue, instead of relying on verbose natural language descriptions, we proposed field-based strategy serializes each check-in into a concise sequence of spatial-temporal attributes separated by vertical bars. 
% As shown in Table~\ref{tab:trajectory_prompt}, this prompt is organized into four functional parts: an instruction block, a history block, a current block, and an output block.
% Applying this field-based design, the history block records past visits exactly as \texttt{trajectory\_id | time | day\_of\_week | norm\_in\_day\_time | poi\_category}. 
% For the target check-in, the current block provides the corresponding temporal context alongside the target trajectory identifier, but masks the target category.
% Finally, the output block requires the model to predict the POI category of the target visit based on the provided context. 

% By employing this compact representation, ANCHOR significantly reduces token overhead. 
% More importantly, by focusing on semantic categories and temporal patterns instead of concrete POI identifiers, this design enables the LLM to better infer the user's underlying mobility intent. This is empirically validated in the Model Efficiency section of our experiments.

\definecolor{codebg}{RGB}{248,248,248}
\definecolor{tagcolor}{RGB}{0,112,112}      % 标签 <...> 颜色
\definecolor{phcolor}{RGB}{180,95,6}        % 占位符 [...] 颜色
\definecolor{headerblue}{RGB}{222,235,247}

% 着色命令修改：使用 \textcolor 避免 p 列首行出现空行
\newcommand{\tg}[1]{\textcolor{tagcolor}{\ttfamily\textbf{#1}}}   % 标签
\newcommand{\pl}[1]{\textcolor{phcolor}{\ttfamily #1}}            % 占位符
\newcommand{\kw}[1]{{\ttfamily #1}}                               % 普通代码

\begin{table}[t]
\centering
\caption{Trajectory prompt template for semantic intent pre-estimation.}
\label{tab:trajectory_prompt}
\small
\renewcommand{\arraystretch}{1.45}
\setlength{\tabcolsep}{8pt}
\begin{tabular}{>{\bfseries}p{0.18\linewidth} >{\columncolor{codebg}}p{0.70\linewidth}}
\toprule
\rowcolor{headerblue}
\textbf{Block} & \textbf{Template} \\
\midrule

Instruction & 
You will be given earlier check-ins of user \pl{[user\_id]}. \newline
Your task is to predict the POI category of the target check-in. \\
\midrule

History &
\tg{<history>} \newline
\kw{Format: trajectory\_id | time | day\_of\_week | norm\_in\_day\_time | poi\_category} \newline
\pl{[trajectory\_id | time | day\_of\_week | norm\_in\_day\_time | poi\_category]} \newline
\kw{...} \newline
\tg{</history>} \\
\midrule

Current &
\tg{<current>} \newline
\pl{[target\_trajectory\_id | target\_time | day\_of\_week | norm\_in\_day\_time]} \newline
\tg{</current>} \\
\midrule

Output &
Return the target POI category only. \newline
\kw{Answer:} \\

\bottomrule
\end{tabular}
\end{table}

\noindent\textbf{LLM Category Classifier.}
After converting the user's check-in history into the textualized trajectory prompt $X_u$, ANCHOR formulates mobility intent pre-estimation as a supervised category classification task rather than open-ended text generation, so that the inferred intent can be represented as a stable and comparable semantic label.
Let $\mathcal{C}$ denote the POI category set, and $c_u^*\in\mathcal{C}$ is the ground-truth category of the target check-in for user $u$.
We instantiate the category predictor by attaching a sequence classification head to the base LLM, whose output dimension equals the number of POI categories $|\mathcal{C}|$.
Given the tokenized prompt $X_u$, the model predicts a probability distribution over POI categories:
\begin{equation}
P_{\theta}(c\mid X_u)=\mathrm{softmax}\left(f_{\theta}^{\mathrm{cls}}(X_u)\right)_c,
\quad c\in\mathcal{C},
\end{equation}
where $f_{\theta}^{\mathrm{cls}}(\cdot)$ denotes the LLM equipped with the sequence classification head.
This formulation restricts the LLM output to the predefined POI category space, avoiding the entity mismatch and uncontrollable outputs caused by directly generating concrete POI identifiers.

\noindent\textbf{LoRA Fine-tuning.}
To adapt the LLM to mobility trajectory data, ANCHOR fine-tunes the model on the category prediction task.
Considering full-parameter fine-tuning of large language models is computationally expensive and introduces unnecessary training overhead for this domain adaptation setting ~\cite{li2024large}.
Therefore, ANCHOR adopts Low-Rank Adaptation (LoRA) to efficiently update the LLM while keeping the backbone parameters frozen~\cite{hu2022lora}.
Specifically, the backbone parameters of the base LLM are kept frozen, and LoRA adapters are injected into the attention projection layers and feed-forward projection layers, including the query/key/value/output projections and feed-forward projections.
Meanwhile, the sequence classification head is kept as a task-specific trainable module.
For an adapted linear transformation matrix $\mathbf{W}$, LoRA represents its effective weight as 
\begin{equation}
    \mathbf{W}'=\mathbf{W}+\Delta \mathbf{W}, ~~~\Delta \mathbf{W}=\frac{\alpha}{r}\mathbf{B}\mathbf{A}.
\end{equation}
Here, $\mathbf{W}$ is the frozen original weight matrix, $\mathbf{A}$ and $\mathbf{B}$ are trainable low-rank adaptation matrices, $r$ is the LoRA rank, and $\alpha$ is the scaling factor.
In this way, ANCHOR updates only a small number of adapter and classification head parameters while adapting the base LLM to the trajectory category prediction task.

Given the category prediction training set $\mathcal{D}_{\text{cat}}$, where each sample consists of a textualized trajectory prompt $X_u$ and its ground-truth target category $c_u^*$. 
ANCHOR trains the category classifier with the cross-entropy loss: \begin{equation}
\mathcal{L}_{\text{cat}}
=
-\frac{1}{|\mathcal{D}_{\text{cat}}|}
\sum_{(X_u,c_u^*)\in\mathcal{D}_{\text{cat}}}
\log P(c_u^*\mid X_u).
\end{equation}
Here, $P(c_u^*\mid X_u)$ is the probability of the ground-truth target category obtained by applying softmax to the sequence classification logits.
Minimizing $\mathcal{L}_{\text{cat}}$ encourages the model to assign higher probabilities to the true target POI categories according to the historical category sequence and the target-time context.
\subsection{Multi-Source Candidate Grounding}

Semantic intent pre-estimation yields a stable category-level mobility intent, but such intent cannot be used directly as the final prediction. 
In real-world check-in data, a single POI category usually corresponds to many concrete locations, and POIs under the same category may differ substantially in historical relevance, personalized preference, recent behavior, and target-time context. 
Hence, a grounding ambiguity remains between high-level category intent and concrete POI entities.

A straightforward solution is to recall POIs only from the LLM-predicted categories, but this is not sufficiently reliable. 
When the predicted categories are incomplete, the ground-truth POI may be missed because its category is not covered. 
Conversely, when too many or overly broad categories are considered, the candidate pool is flooded with weakly related POIs, which complicates subsequent reranking. 
Candidate grounding should therefore not be treated as a simple category-to-POI expansion, but should incorporate multiple types of evidence relevant to concrete POI selection.
Inspired by the multi-source candidate generation paradigm in~\cite{covington2016deep}, ANCHOR implements candidate grounding as a multi-source recall process.
This process consists of three key components: candidate source selection,  grounding prior scoring, and quota-controlled candidate selection.

\noindent\textbf{Candidate Source Selection.}
To ground category-level intent into concrete POIs, ANCHOR constructs the candidate space from three complementary sources of evidence rather than a single source, as summarized in Table~\ref{tab:candidate_sources}. 
\begin{itemize}[leftmargin=*] 
    \item \textbf{LLM-category POIs} are the most direct grounding source: they retrieve POIs under the top-\(K_c\) categories predicted by the LLM, keeping candidates consistent with the inferred mobility intent. However, since one category usually contains many POIs, category consistency alone cannot determine which location the user will finally visit.
    
    \item \textbf{History POI memory} keeps recently or repeatedly visited POIs in the candidate space, capturing revisit behavior that is widely observed in human mobility~\cite{cheng2013you,feng2015personalized}.
    
    \item \textbf{User-preferred category expansion} compensates for cases where the LLM prediction is too narrow or incomplete, supplementing candidates according to the user's long-term, recent, and target-time category preferences~\cite{sun2020go,yuan2013time}.
\end{itemize}
In this way, the candidate pool is constrained by semantic intent while remaining sensitive to user-specific mobility evidence.

\begin{table}[t]
\centering
\caption{Multi-source evidence used for candidate grounding.}
\label{tab:candidate_sources}
\small
\renewcommand{\arraystretch}{1.08}
\begin{tabularx}{\columnwidth}{
>{\RaggedRight\arraybackslash}p{0.32\columnwidth}
>{\RaggedRight\arraybackslash}p{0.36\columnwidth}
>{\RaggedRight\arraybackslash}X
}
\toprule
\textbf{Source} & \textbf{Evidence} & \textbf{Role} \\
\midrule
LLM-category POIs
&
POIs under top-$K_c$ categories
&
Semantic grounding
\\
History POI memory
&
Repeated and recent POIs
&
Revisit behavior
\\
User-preferred category expansion
&
Long-term, recent, and target-time preferences
&
Personalized expansion
\\
\bottomrule
\end{tabularx}
\end{table}

\noindent\textbf{Grounding Prior Scoring.}
After determining the candidate sources, ANCHOR assigns a grounding prior score to each recalled candidate POI.
This score measures how strongly the candidate is supported by multi-source evidence during candidate grounding.
It is not the final recommendation score, but it is used to organize and select candidate POIs before reranking.
Specifically, the grounding prior score of candidate $p$ consists of LLM category evidence, user preference evidence, and history memory evidence:
\begin{equation}
 r(p) = r_{\text{llm}}(p) + r_{\text{pref}}(p) + r_{\text{hist}}(p). 
\end{equation}
\begin{itemize}[leftmargin=*]
    \item \textbf{LLM category evidence.} $r_{\text{llm}}(p)$ measures the semantic consistency between candidate p and the mobility intent predicted by the LLM. Let $\mathcal{C}_{\text{cand}}$ denote the set of top-$K_c$ categories predicted by the LLM. 
    If the category $c_p$ of candidate $p$ lies in $\mathcal{C}_{\text{cand}}$, ANCHOR scores it according to the category rank: 
    \begin{equation} 
   r_{\text{llm}}(p)
    =
    w_{\text{llm}}
    \cdot
    \frac{1}{\log(\operatorname{rank}(c_p)+1)},
    \end{equation} 
    where $\operatorname{rank}(c_p)$ is the rank of $c_p$ in the LLM-predicted list. A higher-ranked category reflects greater LLM confidence, so its POIs receive a higher semantic grounding prior.

    \item \textbf{History memory evidence.} $r_{\text{hist}}(p)$ captures concrete location memory from the user's historical visits, considering both the visit frequency and the recency of candidate $p$ in the visible history:
    \begin{equation}
    \begin{aligned}
    r_{\text{hist}}(p)
    = \;& w_{\text{rep}}\log(1+n_u(p)) \\
    & + w_{\text{rec}}\sum_{i:p_i=p}\frac{1}{\log(\Delta_i+1)},
    \end{aligned}
    \end{equation}
    where $n_u(p)$ is the number of times user $u$ has visited $p$, and $\Delta_i$ is the position distance between the $i$-th visit to $p$ and the target time. POIs visited more frequently or more recently receive higher history priors, helping preserve revisit targets.

    \item \textbf{User preference evidence.} $r_{\text{pref}}(p)$ measures how well candidate $p$ matches the user's category preference. 
    ANCHOR estimates three preference signals from the user's history: long-term, recent, and target-time category preferences:
    \begin{equation}
    \begin{aligned}
    r_{\text{pref}}(p)
    = \;& w_{\text{long}}\,\phi_{\text{long}}(c_p)
    + w_{\text{recent}}\,\phi_{\text{recent}}(c_p) \\
    & + w_{\text{time}}\,\phi_{\text{time}}(c_p),
    \end{aligned}
    \end{equation}
    where $\phi_{\text{long}}(c_p)$, $\phi_{\text{recent}}(c_p)$, and $\phi_{\text{time}}(c_p)$ denote the user's long-term, recent, and target-time preference strengths for category $c_p$. This provides additional personalized grounding support to POIs under user-preferred categories.
\end{itemize}
The coefficients $w_{\text{llm}}$, $w_{\text{rep}}$,  $w_{\text{rec}}$, $w_{\text{long}}$, $w_{\text{recent}}$, and $w_{\text{time}}$  are scaling factors that calibrate the evidence terms from different sources into a unified grounding prior. 
When a POI is recalled by multiple sources, ANCHOR preserves its source flags and decomposed evidence features, so that the history-aware reranker can further exploit such multi-source support.

\noindent\textbf{Quota-Controlled Candidate Selection.} Although the grounding prior reflects the evidence strength of each candidate, truncating candidates by a single global score may introduce source bias. 
For example, LLM-category POIs typically form a much larger set and, without additional constraints, could dominate the pool and crowd out smaller but important sources such as history POI memory. 
ANCHOR therefore imposes source-specific quotas when constructing the candidate pool, so that each source retains a number of high-priority candidates.

Concretely, ANCHOR first ranks candidates within each source by their grounding priors and selects the top ones under the corresponding quota. 
The selected candidates from all sources are then merged and deduplicated into an initial pool. 
If the pool size is still below the predefined size $N$, ANCHOR fills the remaining slots with the highest-scored unselected candidates until it reaches $N$ or no candidates remain. 
The resulting candidate pool is denoted $\mathcal{P}_{\mathrm{cand}}$. 
In this way, ANCHOR balances candidate coverage, source complementarity, and reranking efficiency.

\subsection{History-Aware Reranking}
After multi-source candidate grounding, ANCHOR obtains a candidate POI set annotated with source flags, grounding prior scores, and evidence features. 
These signals explain \emph{why} a candidate is recalled, but not whether it should be ranked high under the current context, since recall-stage evidence mainly reflects source-level support and lacks fine-grained matching between the candidate and the user's current state. 
Producing the final prediction still requires evaluating how well each candidate matches the user's historical behavior, target-time condition, and recent mobility state.

\begin{figure}[t]
\centering
\includegraphics[width=\columnwidth]{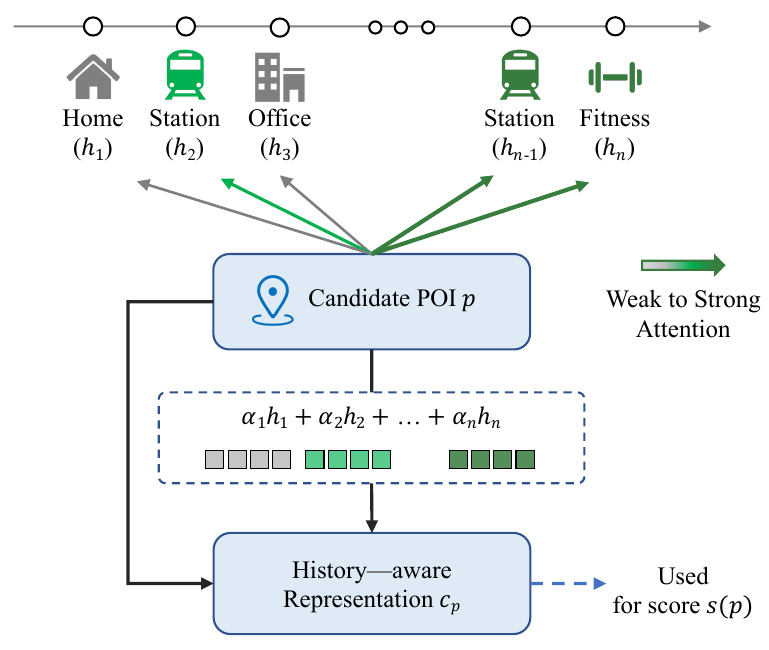}
\caption{Illustration of candidate-specific attention over user history in the history-aware reranking stage.}
\label{fig:history_attention}
\end{figure}

Therefore, ANCHOR introduces a history-aware reranker to re-evaluate the contextual relevance of each candidate within the candidate pool.
This candidate-specific matching is inspired by target attention~\cite{zhou2018deep}, which dynamically activates historical behaviors relevant to each candidate.
As shown in Fig.~\ref{fig:history_attention}, the key idea is to let each candidate POI select the most relevant historical visits and predict the final ranking score based on candidate-specific historical evidence.

\noindent\textbf{History and Candidate Representation.}
To model the matching relationship between candidate POIs and user history, ANCHOR needs to encode historical check-ins and candidate POIs into the same hidden space.
For the $i$-th check-in in the visible history, the model fuses entity, category, temporal, and trajectory-position information into a history representation: 
\begin{equation} \mathbf{h}_i=
    \mathrm{Proj}_h\left( [\mathbf{e}^{\text{poi}}_i;~\mathbf{e}^{\text{cat}}_i;~\mathbf{e}^{\text{time}}_i;~\mathbf{e}^{\text{pos}}_i] \right), 
\end{equation} 
where the four embeddings correspond to the POI, its category, the visiting time, and the trajectory position, and $\mathrm{Proj}_h(\cdot)$ is a linear projection into the hidden space.
The POI embeddings are initialized with LightGCN representations learned from the POI transition graph~\cite{he2020lightgcn}, so that historical and candidate POIs share structured entity semantics.
As a result, stacking all check-ins yields the history sequence: \begin{equation} 
    \mathbf{H}_u = [\mathbf{h}_1,~\mathbf{h}_2, \ldots, \mathbf{h}_n]. 
\end{equation}

Each candidate POI is encoded analogously. For a candidate p, ANCHOR fuses its entity attributes with the candidate-stage evidence features into a candidate representation: 
\begin{equation} 
    \mathbf{v}_p
    =
    \mathrm{Proj}_c
    \left(
    [
    \mathbf{e}^\text{poi}_p;
    \mathbf{e}^\text{cat}_p;
    \mathbf{e}^\text{reg}_p;
    \mathbf{x}_p
    ]
    \right),
\end{equation} 
where the first three embeddings correspond to the candidate POI, its category, and its spatial region, and $x_p$ collects the grounding evidence and context-matching features that characterize why $p$ is recalled and how it initially relates to the current user state.
$\mathrm{Proj}_c(\cdot)$ is the candidate-side linear projection. 
With both sides embedded in a unified space, the reranker can model fine-grained interactions between candidates and historical check-ins, where each candidate attends to its most relevant historical visits (Fig.~\ref{fig:history_attention}).

\noindent\textbf{Current-Context Query Construction.}
Beyond the history and candidate representations, the reranker must also characterize the current prediction state. 
ANCHOR builds three context queries that capture, respectively, the recent historical state, the target-time condition, and the short-term transition state.
First, the model applies position-aware pooling over the history tokens to summarize the user's state before the target time: 
\begin{equation} 
    \mathbf{s}_u = \sum_{i=1}^{n}\beta_i,\mathbf{h}_i, 
\end{equation}
where $\beta_i$ is a pooling weight normalized from the position of the $i$-th check-in and $\mathbf{s}_u$ summarizes the user's historical state before the target time.
Then, ANCHOR constructs three context queries from different contextual signals: the temporal query is built from target day-of-week, target time bin, and time interval; the transition query is built from the previous POI, previous category, and time interval; and the summary query is built from the last history token and the history summary:
\begin{equation}
\begin{aligned}
\mathbf{q}_t
&=
\mathrm{Proj}_t
\left(
[
\mathbf{e}^\text{dow}_{t^*};
\mathbf{e}^\text{bin}_{t^*};
\mathbf{e}^{\Delta}_{t^*}
]
\right), \\
\mathbf{q}_{tr}
&=
\mathrm{Proj}_{tr}
\left(
[
\mathbf{e}^\text{poi}_\text{prev};
\mathbf{e}^\text{cat}_\text{prev};
\mathbf{e}^{\Delta}_{t^*}
]
\right), \\
\mathbf{q}_s
&=
\mathrm{Proj}_s
\left(
[
\mathbf{h}_n;
\mathbf{s}_u
]
\right).
\end{aligned}
\end{equation}
Here, \(\mathrm{Proj}_t\), \(\mathrm{Proj}_{tr}\), and \(\mathrm{Proj}_s\) are learnable linear projection layers that map heterogeneous context features into the same hidden space.
Specifically, \(\mathbf{q}_t\) represents the target-time condition, \(\mathbf{q}_{tr}\) represents the short-term transition state from the previous visit to the target time, and \(\mathbf{q}_s\) represents the recent historical state.

Since the relative importance of these queries varies across cases, ANCHOR does not concatenate them into a static context vector. Instead, it first models their dependencies with a lightweight self-attention encoder~\cite{vaswani2017attention}:
\begin{equation}
[
\hat{\mathbf{q}}_t;
\hat{\mathbf{q}}_{tr};
\hat{\mathbf{q}}_s
]
=
\operatorname{SelfAttn}
\left(
[
\mathbf{q}_t;
\mathbf{q}_{tr};
\mathbf{q}_s
]
\right).
\end{equation}
Then, ANCHOR uses $\mathbf{q}_s$ as the pooling query to perform attention pooling over the encoded context queries:
\begin{equation}
\mathbf{q}
=
\operatorname{AttnPool}
\left(
\mathbf{q}_s,
[
\hat{\mathbf{q}}_t;
\hat{\mathbf{q}}_{tr};
\hat{\mathbf{q}}_s
]
\right).
\end{equation}
We adopt $\mathbf{q}_s$ as the pooling query because, being derived from the user's recent historical state, it serves as a natural anchor for adaptively aggregating target-time and transition information. 
The resulting $\mathbf{q}$ represents the overall prediction state and is used to match it with candidate POIs.

\noindent\textbf{Candidate-History Attention.}
After obtaining the history representation sequence $\mathbf{H}_u$ and the candidate representation $\mathbf{v}_p$, ANCHOR models how each candidate matches the user's historical records, letting the candidate directly interact with history tokens and attend to its most relevant visits.
Using the candidate representation as the query and the history tokens as keys and values, the attention weight of candidate $p$ on the $i$-th check-in is:
\begin{equation}
\alpha_{p,i}
=
\operatorname{softmax}_i
\left(
\frac{
\mathbf{v}_p^{\top}\mathbf{h}_i
}
{\sqrt{d}}
\right),
\end{equation}
where $\alpha_{p,i}$ indicates how important check-in $i$ is for evaluating candidate $p$. 
The history tokens are then aggregated into a candidate-specific historical context: 
\begin{equation}
\mathbf{c}_p
=
\sum_{i=1}^{n}
\alpha_{p,i}\mathbf{h}_i .
\end{equation}

To combine candidate-side information with this historical evidence, ANCHOR fuses the candidate representation, the historical context, and their element-wise interaction:
\begin{equation}
\tilde{\mathbf{v}}_p
=
\operatorname{LN}
\left(
\mathbf{v}_p
+
\operatorname{MLP}
\left(
[
\mathbf{v}_p;
\mathbf{c}_p;
\mathbf{v}_p\odot\mathbf{c}_p
]
\right)
\right).
\end{equation}
Here, $\odot$ denotes element-wise product, $\operatorname{MLP}(\cdot)$ fuses the candidate representation, historical context, and their matching relationship, and $\operatorname{LN}$ denotes layer normalization.
The resulting $\tilde{\mathbf{v}}_p$ is a history-aware candidate representation that captures the fine-grained matching relationship between candidate $p$ and the user's historical visiting patterns.

\noindent\textbf{Context-Aware Scoring.}
Finally, ANCHOR matches the history-aware representation against the current prediction state to produce the ranking score.
For candidate $p$, the model computes element-wise interactions between $\tilde{\mathbf{v}}_p$ and the overall context representation $\mathbf{q}$, the target-time context $\mathbf{q}_t$, and the short-term transition context $\mathbf{q}_{tr}$ and concatenates them with $\tilde{\mathbf{v}}_p$:
\begin{equation}
\mathbf{z}_p
=
[
\tilde{\mathbf{v}}_p \odot \mathbf{q};
\tilde{\mathbf{v}}_p \odot \mathbf{q}_t;
\tilde{\mathbf{v}}_p \odot \mathbf{q}_{tr};
\tilde{\mathbf{v}}_p
].
\end{equation}
The three interaction terms capture how candidate p matches the overall prediction state, the target-time condition, and the short-term transition state, respectively. 
An MLP scorer then maps $\mathbf{z}_p$ to a ranking score:
\begin{equation}
s(p)
=
\operatorname{MLP}_s(\mathbf{z}_p).
\end{equation}
The score $s(p)$ determines the relative priority of POIs within the candidate pool.
During training, ANCHOR applies a softmax over all candidate scores and minimizes the cross-entropy loss against the ground-truth 
\begin{equation}
\mathcal{L}_{\text{rank}}
=
-\log
\frac{
\exp(s(p^*))
}
{
\sum_{p'\in\mathcal{P}_\text{cand}}\exp(s(p'))
}.
\end{equation}
At inference, ANCHOR ranks the candidates by $s(p)$ and returns the top-scored POI as the final recommendation. 
In this way, the reranker integrates candidate-specific historical evidence, the target-time condition, and the short-term transition state into a unified scoring process.

\section{Experiments}

\subsection{Experimental Settings}

\noindent\textbf{Datasets.}
We conduct experiments on three public real-world check-in datasets, including Foursquare-NYC, Foursquare-TKY, and Gowalla-CA, as summarized in Table~\ref{tab:dataset}.
Foursquare-NYC contains check-ins collected in New York City from April 2012 to February 2013, while Foursquare-TKY was collected in Tokyo during the same period~\cite{yang2014modeling}.
Gowalla-CA contains check-ins posted on the Gowalla platform in California and Nevada from February 2009 to October 2010~\cite{yuan2013time}.
Each record contains user information, a point-of-interest (POI), the corresponding POI category, GPS coordinates, and a timestamp.
Following the common preprocessing protocol in prior next POI recommendation studies~\cite{yan2023spatio}, we remove POIs with fewer than 10 check-ins and users with fewer than 10 check-ins.
We then split each dataset chronologically into training, validation, and test sets at 80\%, 10\%, and 10\% ratios strictly at the trajectory level, with each trajectory serving as an indivisible unit.

\begin{table}[htbp]
    \centering
    \caption{Statistics of the datasets}
    \label{tab:dataset}
    \begin{tabular}{lccccc}
        \toprule
        \textbf{Dataset} & \textbf{\#user} & \textbf{\#poi} & \textbf{\#cat} & \textbf{\#checkin} & \textbf{\#trajectory} \\
        \midrule
        NYC & 1,075 & 5,099 & 318 & 104,074 & 14,160 \\
        TKY & 2,281 & 7,844 & 291 & 361,430 & 44,692 \\
        CA  & 4,318 & 9,923 & 301 & 250,780 & 32,920 \\
        \bottomrule
    \end{tabular}
\end{table}

\noindent\textbf{Evaluation Metrics.}
We adopt Acc@K and mean reciprocal rank (MRR) as evaluation metrics for next POI recommendation.
For each test sample, the model outputs a ranked list of POIs.
Acc@K measures whether the ground-truth POI appears in the top-$K$ positions of the ranked list:
\begin{equation}
% \mathrm{Acc@K}
\mathrm{Acc}@K
=
\frac{1}{|\mathcal{D}_{\text{test}}|}
\sum_{q\in\mathcal{D}_{\text{test}}}
\mathbb{I}(r_q \le K),
\end{equation}
where $\mathcal{D}_{\text{test}}$ denotes the test set, $r_q$ denotes the rank position of the ground-truth POI for sample $q$, and $\mathbb{I}(\cdot)$ is the indicator function.
We report Acc@1, Acc@5, and Acc@10 to evaluate recommendation accuracy under different top-$K$ settings.

MRR evaluates the overall ranking quality by considering the reciprocal rank of the ground-truth POI:
\begin{equation}
\mathrm{MRR}
=
\frac{1}{|\mathcal{D}_{\text{test}}|}
\sum_{q\in\mathcal{D}_{\text{test}}}
\frac{1}{r_q}.
\end{equation}
A higher MRR indicates that the ground-truth POI is ranked closer to the top of the recommendation list.

\noindent\textbf{Baselines.}
We compare ANCHOR with the following representative next POI recommendation baselines, covering sequential, spatio-temporal, structure-aware, temporal, and LLM-based methods.
\begin{itemize}
    \item \textbf{PRME}~\cite{feng2015personalized}: A personalized ranking metric embedding method that models sequential POI transitions and user-POI preferences in a latent metric space.
    \item \textbf{LSTM}~\cite{liu2016predicting}: A recurrent sequential model that captures dependency patterns from historical check-in sequences.
    \item \textbf{PLSPL}~\cite{he2016inferring}: A personalized next POI recommendation model that exploits latent behavior patterns and combines long- and short-term mobility preferences.
    \item \textbf{STAN}~\cite{luo2021stan}: A spatio-temporal attention network that models correlations among visits within user trajectories.
    \item \textbf{GETNext}~\cite{yang2022getnext}: A transformer-based method that uses trajectory flow maps to enhance next POI prediction with collaborative mobility signals.
    \item \textbf{STHGCN}~\cite{yan2023spatio}: A spatio-temporal hypergraph learning method that captures higher-order relations in check-in data.
    \item \textbf{ROTAN}~\cite{feng2024rotan}: A temporal next POI recommendation method that models time-specific transition patterns with rotation-based temporal attention.
    \item \textbf{LLM4POI}~\cite{li2024large}: An LLM-based method that converts mobility data into language-readable prompts for next POI recommendation.
    \item \textbf{GNPR}~\cite{wang2025generative}: A generative next POI recommendation method that represents POIs with semantic IDs.
    \item \textbf{GA-LLM}~\cite{liu2025geography}: An LLM-based POI recommendation method that incorporates geographical awareness into mobility prediction.
    \item \textbf{Think2Go}~\cite{zhuang2026think2go}: A reasoning-enhanced generative method that uses LLM reasoning for next POI recommendation.
\end{itemize}

\noindent\textbf{Implementation Details.}
We employ Qwen2.5-7B-Instruct as the base model and fine-tune it using LoRA~\cite{hu2022lora} with rank \(r=16\) and dropout rate 0.05.
The maximum history length is 50 check-ins, and the maximum sequence length is 2560 tokens.
The classifier is trained for 5 epochs with a learning rate of \(5\times10^{-5}\), a batch size of 4, and 4 gradient accumulation steps.
For candidate grounding, we use the top-10 predicted categories and set the candidate pool size to 2000.
The quotas of history POI memory, LLM category anchor, and user preference expansion are set to 100, 800, and 200, respectively.
The remaining slots are filled by candidates from the same enabled sources.
The history-aware reranker uses POI embeddings initialized by LightGCN~\cite{he2020lightgcn} and a two-layer Transformer with hidden size 256.
It is trained with AdamW using a learning rate of \(1\times10^{-5}\), weight decay of 0.02, and early stopping patience of 10.
All experiments are conducted on four NVIDIA GeForce RTX 4090 GPUs.

\begin{table*}[t]
\centering
\caption{Main results on three public next POI recommendation datasets.}
\label{tab:main_results}
\small
\resizebox{\textwidth}{!}{
\begin{tabular}{lcccccccccccc}
\toprule
\multirow{2}{*}{\textbf{Model}}
&
\multicolumn{4}{c}{\textbf{NYC}}
&
\multicolumn{4}{c}{\textbf{TKY}}
&
\multicolumn{4}{c}{\textbf{CA}}
\\
\cmidrule(lr){2-5}
\cmidrule(lr){6-9}
\cmidrule(lr){10-13}
&
\textbf{Acc@1}
&
\textbf{Acc@5}
&
\textbf{Acc@10}
&
\textbf{MRR}
&
\textbf{Acc@1}
&
\textbf{Acc@5}
&
\textbf{Acc@10}
&
\textbf{MRR}
&
\textbf{Acc@1}
&
\textbf{Acc@5}
&
\textbf{Acc@10}
&
\textbf{MRR}
\\
\midrule
PRME
&
0.1159
&
0.2236
&
0.3105
&
0.1712
&
0.1052
&
0.2278
&
0.2944
&
0.1786
&
0.0521
&
0.1034
&
0.1425
&
0.1002
\\
LSTM
&
0.1305
&
0.2719
&
0.3283
&
0.1857
&
0.1335
&
0.2728
&
0.3277
&
0.1834
&
0.0665
&
0.1306
&
0.1784
&
0.1201
\\
PLSPL
&
0.1917
&
0.3678
&
0.4523
&
0.2806
&
0.1889
&
0.3523
&
0.4150
&
0.2542
&
0.1072
&
0.2278
&
0.2995
&
0.1847
\\
STAN
&
0.2231
&
0.4582
&
0.5734
&
0.3253
&
0.1963
&
0.3798
&
0.4464
&
0.2852
&
0.1104
&
0.2348
&
0.3018
&
0.1869
\\
GETNext
&
0.2406
&
0.4815
&
0.5811
&
0.3528
&
0.1829
&
0.4045
&
0.4961
&
0.2853
&
0.1526
&
0.3278
&
0.3946
&
0.2364
\\
STHGCN
&
0.2734
&
0.5361
&
0.6244
&
0.3915
&
0.2950
&
0.5207
&
0.5980
&
0.3986
&
0.1730
&
0.3529
&
0.4191
&
0.2558
\\
ROTAN
&
0.3106
&
0.5281
&
0.6131
&
0.4104
&
0.2458
&
0.4626
&
0.5392
&
0.3475
&
0.2199
&
0.3718
&
0.4334
&
0.2931
\\
LLM4POI
&
0.3372
&
-
&
-
&
-
&
0.3035
&
-
&
-
&
-
&
0.2065
&
-
&
-
&
-
\\
GNPR
&
0.3618
&
-
&
-
&
-
&
0.3062
&
-
&
-
&
-
&
0.2403
&
-
&
-
&
-
\\
GA-LLM
&
0.3919
&
-
&
-
&
-
&
0.3482
&
-
&
-
&
-
&
0.2566
&
-
&
-
&
-
\\
Think2Go
&
0.3847
&
0.6287
&
0.6835
&
0.4663
&
0.3239
&
0.5461
&
0.6051
&
0.4147
&
0.2583
&
0.4190
&
0.4682
&
0.3167
\\
\textbf{ANCHOR}
&
\textbf{0.4138}
&
\textbf{0.6889}
&
\textbf{0.7541}
&
\textbf{0.5335}
&
\textbf{0.3602}
&
\textbf{0.5939}
&
\textbf{0.6692}
&
\textbf{0.4672}
&
\textbf{0.2633}
&
\textbf{0.4328}
&
\textbf{0.4931}
&
\textbf{0.3432}
\\
\bottomrule
\end{tabular}
}
\end{table*}

\subsection{Main Results}

Table~\ref{tab:main_results} reports the overall performance of ANCHOR and all baselines on the three datasets.
Overall, ANCHOR achieves the best performance on NYC, TKY, and CA, and consistently improves Acc@1, Acc@5, Acc@10, and MRR.
This indicates that the advantage of ANCHOR is not limited to a specific dataset or a single evaluation metric, but generalizes across different cities, platforms, and recommendation list lengths.

Compared with non-LLM methods, ANCHOR achieves relative improvements of 33.23\%, 46.54\%, and 19.74\% over the strong baseline ROTAN on Acc@1 for NYC, TKY, and CA, respectively.
Although existing structure-enhanced methods improve POI-level transition modeling through trajectory flow, spatio-temporal graphs, or time-aware mechanisms, their prediction process still mainly operates in the concrete POI space.
Therefore, when the POI space is sparse, long-tailed, and lacks stable semantic representations, relying only on POI-level transition patterns is still insufficient to fully capture the high-level mobility intent behind the user's next visit.
The consistent improvements over these methods show that using the LLM-predicted category intent as a semantic anchor and further grounding it into concrete POI candidates can more effectively bridge the granularity gap between semantic intent and POI entities.

Compared with LLM-based methods, ANCHOR achieves relative improvements of 7.56\%, 11.21\%, and 1.94\% over the strongest baseline Think2Go on Acc@1 for NYC, TKY, and CA, respectively.
More notably, ANCHOR also obtains clear improvements on MRR, outperforming Think2Go by 14.41\%, 12.66\%, and 8.37\% on NYC, TKY, and CA, respectively.
This indicates that ANCHOR not only predicts the top-1 POI more accurately, but also ranks the ground-truth POI more consistently near the top of the recommendation list.
Although existing LLM-based methods have shown that semantic understanding and reasoning abilities are useful for next POI recommendation, directly generating POI IDs or relying on generative POI representations can still suffer from entity mismatch, sparse supervision, and localized long-tailed POI spaces.
In contrast, ANCHOR uses the LLM for more stable category-level intent estimation, and then performs concrete POI recommendation through multi-source candidate grounding and history-aware reranking.
These results further verify that, for next POI recommendation, a more reliable path is not to let the LLM directly predict sparse POI entities, but to first understand high-level mobility intent and then structurally ground it into sortable candidate POIs.

\subsection{Ablation Study}

To further evaluate the contribution of each core component in ANCHOR, we conduct an ablation study on the three datasets by comparing the full framework with four variants:
(1) \textbf{w/o LightGCN}: replaces the LightGCN-initialized POI embeddings with randomly initialized ones, ignoring the structural transition signals from the POI graph.
(2) \textbf{w/o LLM Anchor}: removes the semantic anchor derived from the LLM-predicted categories. Here, candidate recall relies solely on historical visits and personalized preference expansion.
(3) \textbf{w/o Multi-source}: restricts candidate generation to only the LLM category anchor, omitting the supplementary history POI memory and user preference signals.
(4) \textbf{w/o Reranker}: replaces the history-aware reranker with a standard MLP. This variant ranks candidates based on basic feature concatenation without modeling fine-grained candidate-history interactions.

Table~\ref{tab:ablation} presents the Acc@1 results. We observe that the complete ANCHOR framework consistently outperforms all ablation variants across NYC, TKY, and CA, confirming that all four modules jointly contribute to the overall performance:
(1) The performance drop of \textbf{w/o LightGCN} highlights the benefit of incorporating structural transition patterns into POI representations prior to contextual reranking.
(2) The decline in \textbf{w/o LLM Anchor} confirms that LLM-predicted semantic intents offer crucial constraints, preventing the model from overly relying on historical repetition.
(3) The underperformance of \textbf{w/o Multi-source} indicates that an LLM semantic anchor alone is insufficient; integrating history and preference signals significantly improves the coverage and robustness of the candidate pool.
(4) Finally, the degradation of \textbf{w/o Reranker} demonstrates that merely recalling candidates is inadequate. A careful discrimination through candidate-history attention is vital for identifying the most suitable target POI under the current context.

\begin{table}[t]
\centering
\caption{Results of ablation models on NYC, TKY, and CA datasets with Acc@1 metric.}
\label{tab:ablation}
\small
\setlength{\tabcolsep}{3pt}
\renewcommand{\arraystretch}{1.1}
\resizebox{\columnwidth}{!}{
\begin{tabular}{lccc}
\toprule
\textbf{Ablations} & \textbf{NYC} & \textbf{TKY} & \textbf{CA} \\
\midrule
ANCHOR
&
\textbf{0.4137}
&
\textbf{0.3602}
&
\textbf{0.2633}
\\
w/o LightGCN
&
0.3875 ($\downarrow$6.33\%)
&
0.3341 ($\downarrow$7.25\%)
&
0.2581 ($\downarrow$1.97\%)
\\
w/o LLM Anchor
&
0.3860 ($\downarrow$6.70\%)
&
0.3462 ($\downarrow$3.89\%)
&
0.2569 ($\downarrow$2.43\%)
\\
w/o Multi-source
&
0.3995 ($\downarrow$3.43\%)
&
0.3593 ($\downarrow$0.25\%)
&
0.2540 ($\downarrow$3.53\%)
\\
w/o Reranker
&
0.4002 ($\downarrow$3.26\%)
&
0.3169 ($\downarrow$12.02\%)
&
0.2465 ($\downarrow$6.38\%)
\\
\bottomrule
\end{tabular}
}
\end{table}

\subsection{Effectiveness of Semantic Intent Pre-estimation}

\begin{figure}[t]
\centering
\includegraphics[width=\columnwidth]{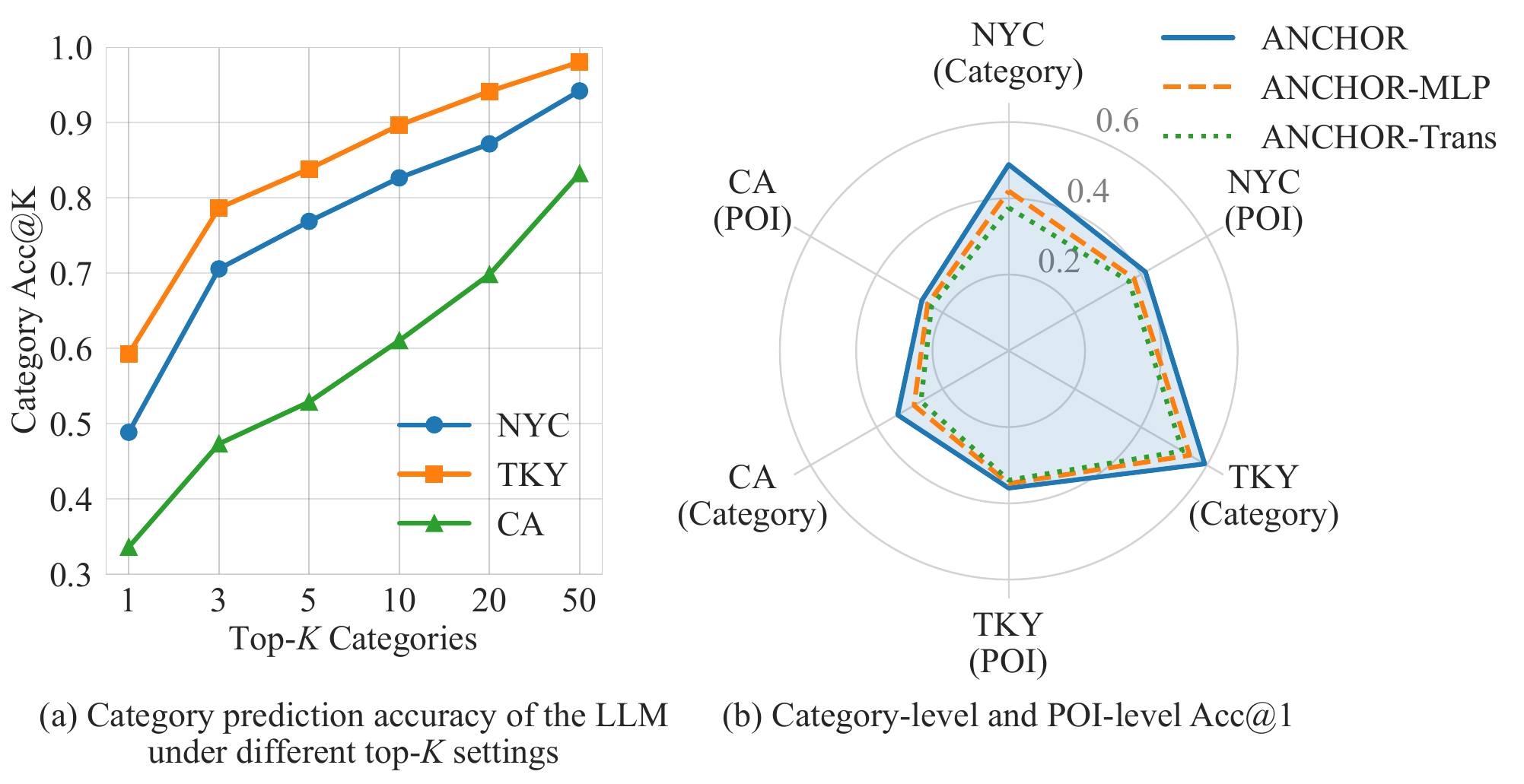}
\caption{Effectiveness of semantic intent pre-estimation.}
% (a) Category prediction accuracy of the LLM under different top-$K$ settings.
% (b) Category-level and POI-level Acc@1 under different category predictors.}
\label{fig:semantic_intent_effectiveness}
\end{figure}

To verify whether ANCHOR can effectively extract the user's high-level mobility intent, we analyze semantic intent pre-estimation from two perspectives.
First, we evaluate whether the top-$K$ categories predicted by the LLM can cover the ground-truth category of the target POI, which measures the recall ability of category-level intent estimation.
Second, we replace the fine-tuned LLM with an MLP and a Transformer category predictor, and compare both category-level Acc@1 and final POI-level Acc@1.
This allows us to examine whether the LLM is necessary for semantic intent extraction rather than merely serving as a replaceable category classifier.

As shown in Fig.~\ref{fig:semantic_intent_effectiveness}(a), Category Acc@K consistently increases on all three datasets as $K$ becomes larger.
This indicates that the LLM-predicted category set can progressively cover the user's true next-visit intent.
Under the top-10 setting, the Category Acc@K reaches 0.8267, 0.8967, and 0.6105 on NYC, TKY, and CA, respectively.
When $K=50$, the values further increase to 0.9422, 0.9806, and 0.8324.
These results show that the LLM can capture reliable category-level mobility intent from historical visit categories and target-time context, providing effective semantic anchors for subsequent candidate grounding.

Fig.~\ref{fig:semantic_intent_effectiveness}(b) further compares different category predictors.
ANCHOR consistently outperforms ANCHOR-MLP and ANCHOR-Trans on category-level Acc@1 across all three datasets, showing that the fine-tuned LLM is more effective at inferring the user's next semantic intent from textualized trajectories.
More importantly, this advantage also transfers to the final POI recommendation results.
ANCHOR achieves the best POI-level Acc@1 on all datasets, indicating that more accurate category-level intent estimation provides stronger semantic constraints for candidate grounding and improves final reranking.

Overall, these results suggest that the LLM in ANCHOR is not simply a replaceable category classifier.
It plays the role of semantic intent extraction by first understanding the user's high-level mobility intent in the category space and then passing this intent to the grounding and reranking stages.

\subsection{User Cold-start Analysis}
To analyze model performance under different user activity levels, we divide test users into Inactive, Normal, and Very active groups according to the number of historical interactions.
We compare ANCHOR with GA-LLM, the strongest baseline in terms of overall Acc@1.
Table~\ref{tab:user_cold_start} reports the Acc@1 results on different user groups.

\begin{table}[t]
\centering
\caption{User cold-start analysis on NYC, TKY, and CA datasets with Acc@1.}
\label{tab:user_cold_start}
\small
\renewcommand{\arraystretch}{1.12}
\begin{tabular}{llccc}
\toprule
\textbf{User group} & \textbf{Model} & \textbf{NYC} & \textbf{TKY} & \textbf{CA} \\
\midrule
Inactive & GA-LLM & 0.3365 & 0.3276 & 0.2655 \\
Normal & GA-LLM & 0.4416 & 0.3513 & 0.2498 \\
Very active & GA-LLM & 0.4129 & 0.3483 & 0.2437 \\
\midrule
Inactive & ANCHOR & 0.3270 & 0.2879 & 0.2872 \\
Normal & ANCHOR & 0.4750 & 0.3607 & 0.2573 \\
Very active & ANCHOR & 0.4263 & 0.4089 & 0.2592 \\
\bottomrule
\end{tabular}
\end{table}

As shown in Table~\ref{tab:user_cold_start}, ANCHOR generally outperforms GA-LLM on Normal and Very active users, indicating that ANCHOR can better exploit personalized mobility evidence when sufficient historical behavior is available.
In particular, on Very active users, ANCHOR achieves clear improvements on TKY and CA, suggesting that the history-aware reranker benefits from richer historical visits when modeling the matching relationship between candidate POIs and user behavior.

For Inactive users, the results are more dataset-dependent.
ANCHOR outperforms GA-LLM on CA, but underperforms GA-LLM on NYC and TKY.
This indicates that under extreme cold-start scenarios, limited historical behavior may restrict the advantage of history-aware reranking.
Overall, these results show that ANCHOR is effective at extracting personalized evidence from available history, while extremely inactive users remain a challenging scenario for next POI recommendation.

\subsection{Model Efficiency}\label{sec:exp_efficiency}

\begin{figure}[t]
\centering
\includegraphics[width=\columnwidth]{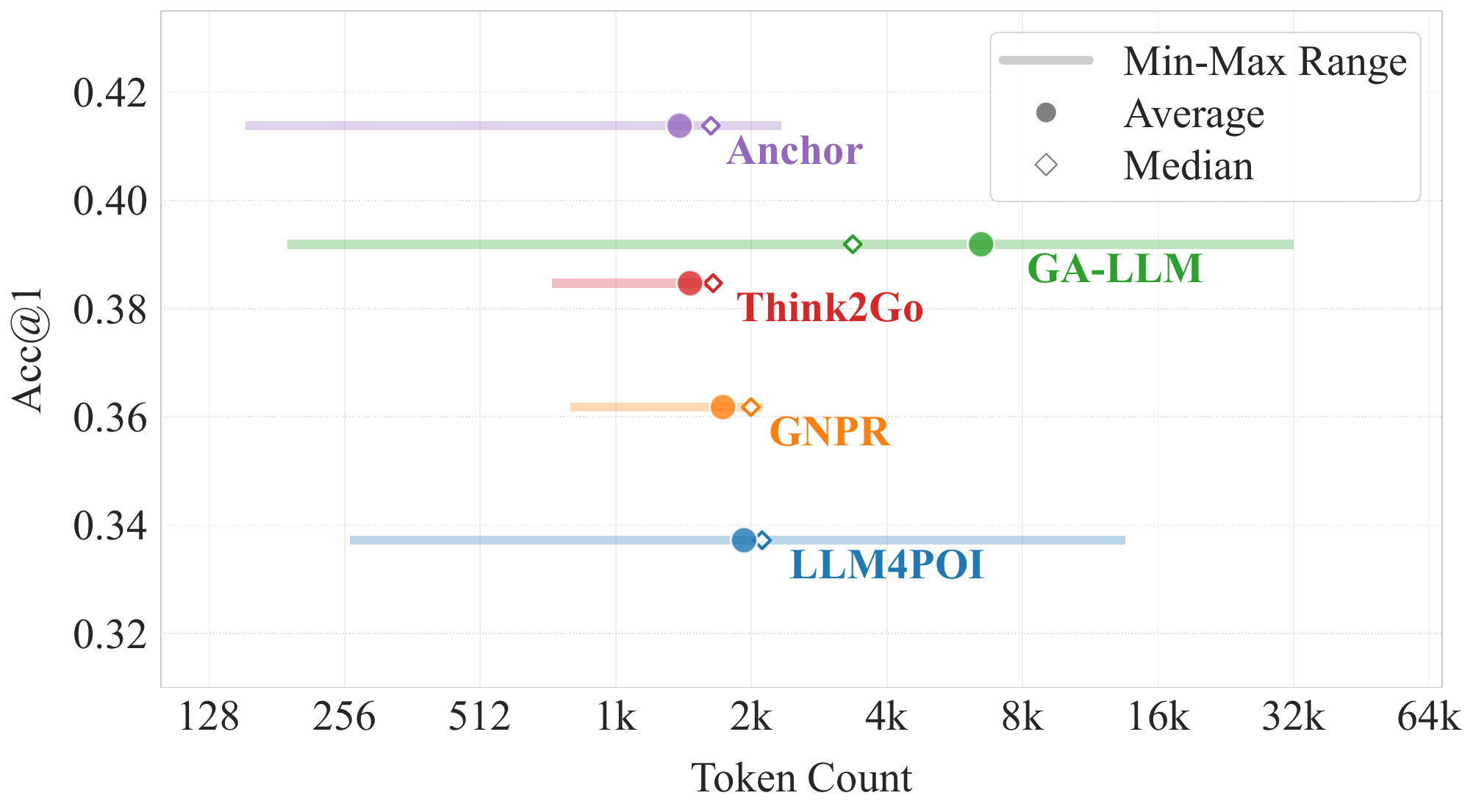}
\caption{Model efficiency comparison on the NYC dataset.
Lines denote min-max prompt token ranges, circles denote average token counts, and diamonds denote median token counts.
The y-axis reports Acc@1.}
\label{fig:model_efficiency}
\end{figure}

LLM-based next POI recommendation is strongly affected by prompt length, as longer inputs increase computation costs and limit the amount of historical behavior that can fit within the context window. 
Therefore, we evaluate the prompt token usage of various LLM-based methods on the NYC dataset to examine their relationship with Acc@1.

As shown in Fig.~\ref{fig:model_efficiency}, ANCHOR is situated in the upper-left region, achieving the highest Acc@1 with a relatively low prompt token cost. 
In contrast, competing methods either consume substantially more tokens without accuracy gains or yield lower Acc@1 despite having similar prompt lengths, demonstrating that ANCHOR achieves a superior trade-off between input cost and recommendation effectiveness.

This efficiency primarily stems from two core designs: our field-based trajectory prompt, which eliminates redundant sentence-level textualization, and our category-level intent estimation, which avoids appending lengthy POI sequences by reasoning in a more compact semantic space.

\subsection{Parameter Sensitivity}

\begin{figure}[t]
\centering
\includegraphics[width=\columnwidth]{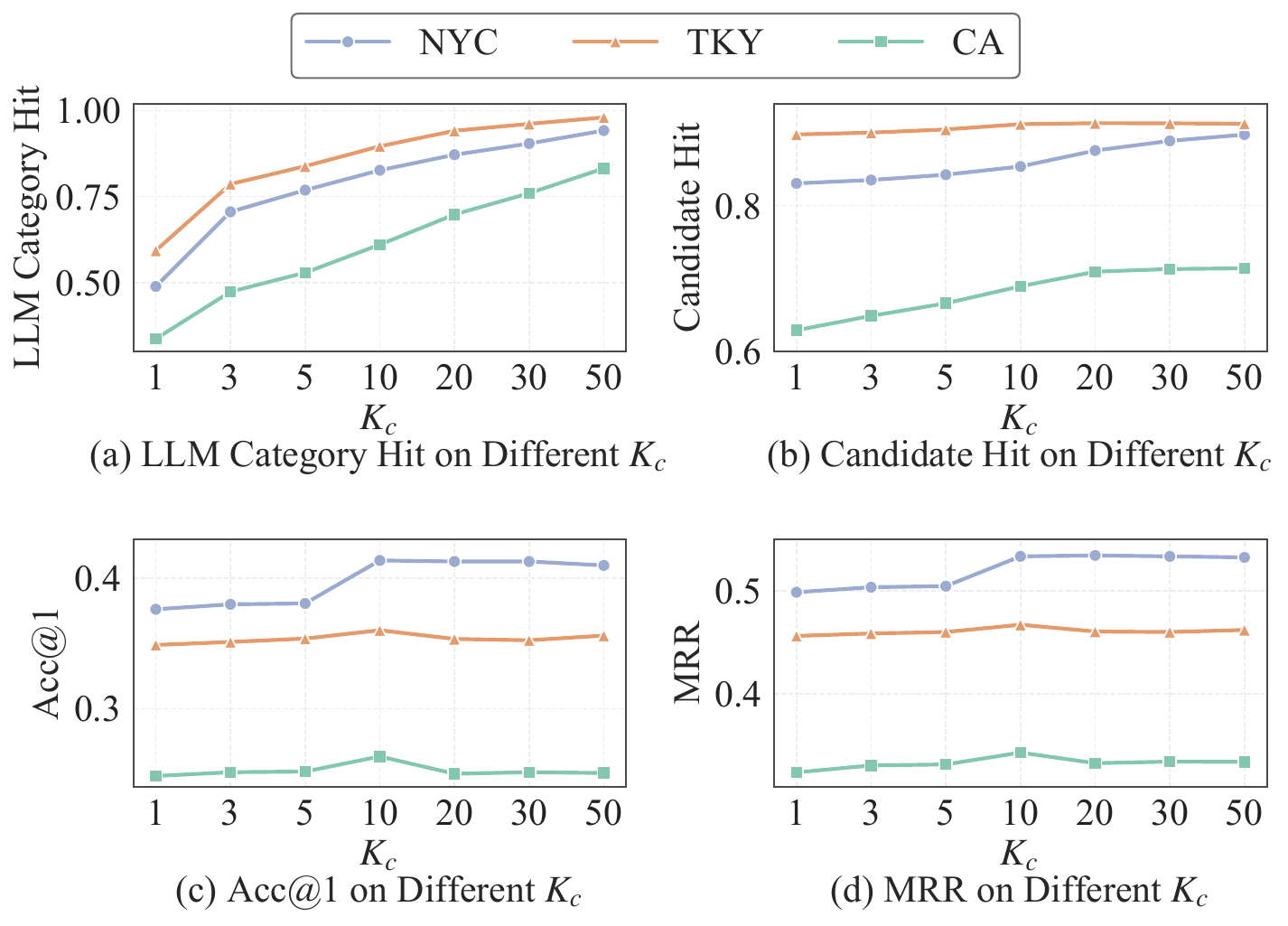}
\caption{Impact of the Number of LLM-Predicted Categories \(K_c\).}
\label{fig:topkc_sensitivity}
\end{figure}

\noindent\textbf{Impact of the Number of LLM-Predicted Categories \(K_c\).}
Fig.~\ref{fig:topkc_sensitivity} shows how coverage-oriented metrics and final recommendation metrics change when different numbers of LLM-predicted categories are used in candidate grounding.
A direct observation is that LLM Category Hit and Candidate Hit generally keep increasing as \(K_c\) becomes larger across the three datasets.
This indicates that including more LLM-predicted categories can better cover the ground-truth category and further increase the chance that the target POI enters the candidate pool.
However, Acc@1 and MRR do not follow the same monotonic trend.
They improve when \(K_c\) increases from small values, but gradually become stable or slightly decrease when \(K_c\) continues to grow.
This suggests that higher candidate coverage does not necessarily lead to better final ranking performance.

This phenomenon can be understood from the relationship between semantic coverage and candidate noise.
A smaller \(K_c\) keeps the candidate pool concentrated around the categories that the LLM is most confident about, leading to stronger semantic constraints but higher sensitivity to category prediction errors.
If the ground-truth category is not included in the top-\(K_c\) categories, the target POI can hardly be recalled.
In contrast, a larger \(K_c\) covers more possible mobility intents and improves category-level and candidate-level coverage.
However, it also introduces more weakly related candidates and weakens the semantic selectivity of category anchors.
Once the candidate pool has covered enough plausible categories, further increasing \(K_c\) mainly adds noisy candidates for the reranker rather than useful candidates.

Therefore, candidate grounding should not simply maximize the coverage probability of the ground-truth POI.
Effective grounding should also preserve the semantic selectivity of the candidate pool, allowing the reranker to distinguish candidates within a relatively high-quality candidate set.
In other words, semantic-to-entity grounding needs to balance category-level uncertainty tolerance with candidate-level discrimination.

\noindent\textbf{Impact of the Number of Candidate POIs \(N\).}
Besides the number of LLM-predicted categories, the candidate pool size \(N\) is another important parameter that affects both recommendation performance and inference efficiency.
Intuitively, a larger candidate pool can cover more potential POIs and thus increase the probability that the ground-truth POI is recalled.
However, expanding the candidate pool may also introduce more weakly related POIs, making the reranking task more difficult.
Moreover, the reranker needs to encode and score more candidate POIs, which increases inference cost.
Therefore, we analyze the impact of \(N\) on Candidate Hit, Acc@1, MRR, and evaluation speed.

\begin{figure}[t]
\centering
\includegraphics[width=\columnwidth]{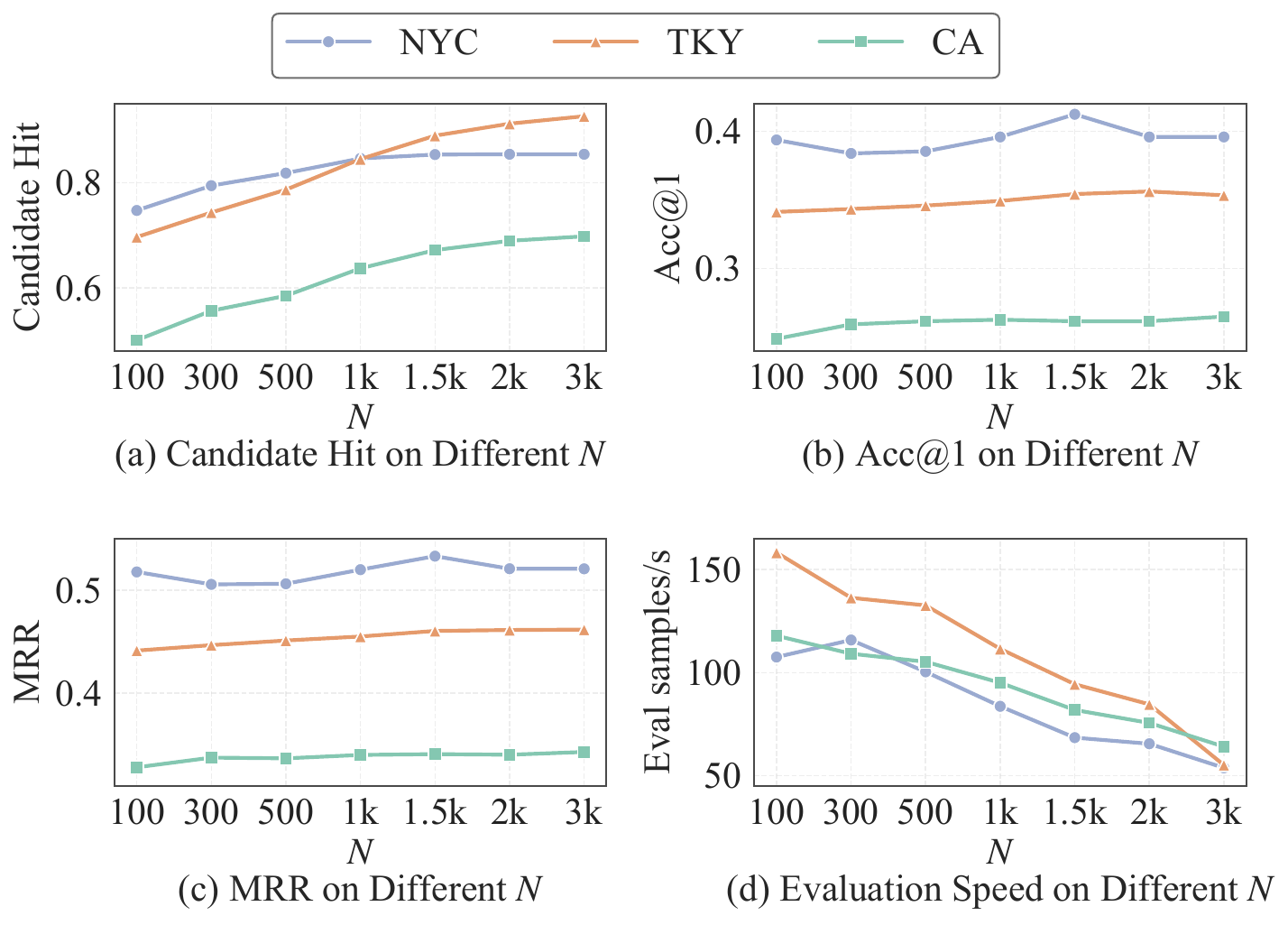}
\caption{Impact of the number of candidate POIs \(N\).}
\label{fig:candidateN_sensitivity}
\end{figure}

To make different candidate pool sizes comparable, we do not keep the original source-specific quotas fixed.
Instead, we scale the quotas of the three recall sources according to the same ratio used in the main experiments.
In this way, candidate pools under different \(N\) values maintain similar multi-source compositions, and the experiment focuses on the effect of candidate pool size itself.
Fig.~\ref{fig:candidateN_sensitivity} shows the performance and efficiency changes when \(N\) increases from 100 to 3000.

As shown in Fig.~\ref{fig:candidateN_sensitivity}, Candidate Hit generally increases on all three datasets as \(N\) becomes larger.
This indicates that expanding the candidate pool indeed improves the probability of including the ground-truth POI in the candidate set.
However, this recall gain does not consistently translate into higher Acc@1.
On NYC, Acc@1 reaches its peak at \(N=1500\), but decreases when the candidate pool is further enlarged.
On TKY and CA, Acc@1 changes more mildly and does not monotonically increase with Candidate Hit.
This suggests that candidate coverage is necessary but not sufficient for improving final recommendation performance.

The evaluation speed shows the opposite trend.
As \(N\) increases, the number of evaluated samples per second drops clearly.
This is because the history-aware reranker needs to model the compatibility between each candidate POI and the user's history and target context.
A larger candidate pool therefore increases the computational cost of representation learning, attention, and scoring.
Thus, an overly large candidate pool can improve candidate coverage, but it may also introduce more ranking noise and reduce inference efficiency.

This experiment shows that candidate grounding needs to balance coverage, ranking quality, and efficiency.
A very small candidate pool may limit final performance due to insufficient recall, while an overly large candidate pool can weaken the reranker's discrimination ability and slow down inference.
Therefore, the candidate pool size is not simply the larger the better.
Effective candidate generation should construct a candidate set that is sufficiently covered yet still compact.

\section{Related Work}

\subsection{Next POI Recommendation}

Next POI recommendation aims to predict the next POI that a user is likely to visit based on historical check-in records.
Early studies usually formulate this task as sequential transition modeling or personalized mobility prediction under spatial, temporal, and preference constraints~\cite{cheng2013you,feng2015personalized,he2016inferring}.
Multi-aspect POI recommendation has also been studied in the data engineering community, showing the relevance of POI prediction to database and mobility-data research~\cite{li2015mars}.
These studies establish localized preference, sequential transition, and latent mobility behavior as fundamental signals for next POI prediction.

With the development of deep learning, recurrent and attention-based models have been widely used to capture spatio-temporal mobility patterns.
HST-LSTM incorporates spatial and temporal factors into LSTM gates for context-aware location transition modeling~\cite{kong2018hst}.
STAN uses spatio-temporal attention to capture correlations among both adjacent and non-adjacent visits within trajectories~\cite{luo2021stan}.
Geography-aware sequential recommendation further introduces geographical influence into sequential location modeling~\cite{lian2020geography}.
ST-PIL models spatial-temporal periodic interest for next POI recommendation~\cite{cui2021st}.
Beyond next POI recommendation, AdaMove studies efficient test-time adaptation for human mobility prediction, further showing the relevance of mobility prediction to data engineering research on spatio-temporal data~\cite{han2025adamove}.
Continual next POI recommendation further studies how models adapt to evolving user interests over time~\cite{wang2025efficient}.

Recently, graph-based and structure-enhanced methods have been introduced to exploit collaborative mobility signals and higher-order spatio-temporal relations.
Seq2Graph augmentation discovers collaborative transition signals from user trajectories~\cite{li2021discovering}.
Graph-based disentangled representation learning separates different mobility factors for next POI recommendation~\cite{wang2022learning}.
GETNext constructs trajectory flow maps to enhance transformer-based next POI recommendation~\cite{yang2022getnext}.
STHGCN builds spatio-temporal hypergraphs to capture higher-order relations~\cite{yan2023spatio}.
AGCL constructs adaptive multi-faceted POI dependency graphs with contrastive learning~\cite{rao2024next}.
ROTAN models time-specific transition patterns with rotation-based temporal attention~\cite{feng2024rotan}.
These methods improve POI-level transition modeling, but they still mainly predict over concrete POI entities and usually lack explicit modeling of high-level mobility intent.

\balance
\subsection{Language Models for Mobility Data}

Large language models have recently been used to introduce semantic knowledge into structured behavioral data.
Prompt-based methods convert user interactions, item descriptions, or sequential behaviors into natural-language inputs for recommendation~\cite{zhang2023prompt,harte2023leveraging}.
Structure-aware methods further combine LLMs with graph signals, collaborative embeddings, or sequential behavior representations.
GraphGPT aligns LLMs with graph-structured data through instruction tuning~\cite{tang2024graphgpt}.
LLMRec augments interaction graphs with LLM-generated side information~\cite{wei2024llmrec}.
CoLLM integrates collaborative embeddings into LLMs for recommendation~\cite{zhang2025collm}.
LLaRA further explores hybrid prompting for sequential recommendation~\cite{liao2024llara}.
These studies provide general mechanisms for injecting semantic information into structured behavior modeling.
However, mobility data introduce additional spatial locality, temporal regularity, and localized entity constraints, which require language models to be adapted specifically for spatio-temporal prediction.

In mobility data modeling, recent studies formulate human mobility forecasting or next POI recommendation as language modeling, semantic representation learning, or generative prediction.
Trajectory-to-language methods translate human mobility forecasting into natural-language generation and further leverage language foundation models for mobility forecasting~\cite{xue2022translating,xue2022leveraging}.
POI-Enhancer uses LLMs to enrich POI representation learning with semantic knowledge~\cite{cheng2025poi}.
LLM4POI adapts LLMs to next POI recommendation through trajectory prompting and trajectory similarity~\cite{li2024large}.
LLM-Move studies the zero-shot generalization ability of LLMs for next POI recommendation~\cite{feng2024move}.
GNPR represents POIs with semantic IDs for generative next POI recommendation~\cite{wang2025generative}.
GA-LLM incorporates geographical awareness into LLM-based POI recommendation~\cite{liu2025geography}.
Think2Go further explores LLM reasoning for generative next POI recommendation~\cite{zhuang2026think2go}.
These methods demonstrate the usefulness of LLMs for modeling semantic and contextual signals in mobility data.
However, textualizing trajectories or generating POI identifiers does not by itself resolve how semantic predictions should be mapped to valid POI entities.

\subsection{Semantic-to-Entity Grounding}

Applying LLMs to structured spatio-temporal data requires mapping high-level semantic predictions back to valid entities in a constrained data space.
Related data engineering studies also show that LLM-based structured prediction often requires grounding model outputs to valid database objects, executable queries, or candidate records~\cite{ren2024purple,fan2025grounding,fan2024cost}.
For next POI recommendation, a category-level intent such as dining, shopping, or commuting may correspond to many candidate POIs.
The valid candidates depend on user history, spatial locality, temporal context, and category compatibility.
Therefore, reliable prediction requires not only semantic reasoning, but also candidate construction and evidence integration over structured mobility records.

Retrieval-enhanced and structure-aware recommendation methods show that external behavioral data, retrieved candidates, and collaborative signals can improve grounding and long-tail prediction~\cite{lin2024rella,wu2024coral}.
However, these methods are not tailored to the localized POI space, where candidate entities are geographically constrained and highly user-dependent.
In LLM-based POI recommendation, directly generating POI identifiers or names may further introduce entity mismatch and scalability issues when the candidate space is large~\cite{li2024large,wang2025generative,zhuang2026think2go}.

ANCHOR follows a semantic-to-entity grounding strategy for next POI recommendation.
It uses the LLM to infer category-level mobility intent, and then grounds this intent into concrete POI candidates through historical memory, category preference, and history-aware reranking.
This design avoids unconstrained POI generation while preserving the semantic reasoning ability of LLMs for structured mobility prediction.

\section{Conclusion}

In this paper, we identify the semantic-to-entity gap in LLM-based next POI recommendation: although LLMs can infer high-level mobility intent from trajectory context, next POI recommendation ultimately requires selecting a concrete POI from a sparse, localized, and long-tailed entity space, creating a granularity mismatch between semantic intent and entity-level prediction.
To bridge this gap, we propose ANCHOR, a category-anchored framework that uses category-level intent as an intermediate semantic representation before grounding it into concrete POI candidates.
This design avoids requiring the LLM to directly generate POI identifiers.
Instead, ANCHOR lets the LLM reason in a more stable category space, grounds the inferred intent into a multi-source candidate POI set with user history and preference evidence, and then performs history-aware reranking for final prediction.
In this way, ANCHOR connects the semantic reasoning ability of LLMs with the entity-level prediction requirement of next POI recommendation.
Experiments on three real-world check-in datasets show that ANCHOR consistently outperforms diverse next POI recommendation baselines.
Further analyses confirm that category-level mobility intent, when grounded with user mobility evidence, can effectively support concrete POI prediction.

A current limitation is that ANCHOR still depends on sufficient historical mobility data for candidate grounding and reranking.
This may limit its robustness for extreme cold-start users with very short check-in histories.
Future work will explore more adaptive grounding strategies for different user contexts and mobility scenarios.
% \section*{Acknowledgment}
% OpenAI ChatGPT was used to assist with language polishing and readability improvement of Section V.
% It was also used to assist in drafting Python scripts for visualizing the experimental results in Fig.~5.
% All scientific content, method design, experimental results, analyses, and final figures were created, reviewed, and verified by the authors.

\section*{Acknowledgement}
This work is supported by the National Natural Science Foundation of China under Grant No. 62506097, U25A20530, U25B2045, by the Shenzhen Basic Research Program Natural Science Foundation under Grant No. JCYJ20250604145542055, and by the Guangdong Basic and Applied Basic Research Foundation under Grant No. 2026A1515010223.

\clearpage

\section*{AI-Generated Content Acknowledgement}
During manuscript preparation, ChatGPT (OpenAI) was used as a language-editing aid for limited English expression and grammar refinement. The research conception, methodology, implementation, evaluation, interpretation of results, and conclusions were developed and verified by the authors.

\bibliographystyle{IEEEtran}
\bibliography{references}

\end{document}